\documentclass[11pt]{article}

\usepackage[final]{acl}
\usepackage[table]{xcolor}
\usepackage{times}
\usepackage{latexsym}

\usepackage[T1]{fontenc}

\usepackage[utf8]{inputenc}

\usepackage{microtype}

\usepackage{inconsolata}
\usepackage{hyperref}
\usepackage{fontawesome5}

\usepackage{graphicx}

\usepackage{subcaption}
\usepackage{rotating}
\usepackage{tcolorbox}
\tcbuselibrary{breakable}
\usepackage{multirow}
\usepackage{makecell}
\usepackage{booktabs}       
\usepackage{hyperref}       
\usepackage{amsmath}
\usepackage{amssymb}
\usepackage{mathtools}
\usepackage{algorithmic}
\usepackage{algorithm}
\usepackage{arydshln}
\usepackage{amsthm}
\usepackage{natbib}
\usepackage{adjustbox}
\usepackage{framed}
\usepackage{wrapfig}

\usepackage[table]{xcolor}
\newcommand{\jygl}{\cellcolor{gray!15}}

\title{Locality-Aware Redundancy Pruning for LLM Depth Compression}

\author{
Vincent-Daniel Yun\textsuperscript{1,2}$^{,\dag}$,
Youngrae Kim\textsuperscript{1}$^{,\dag}$,
Woosang Lim\textsuperscript{2,3}$^{,\dag}$\\
\textbf{Youngjin Heo\textsuperscript{2}, Minkyu Kim\textsuperscript{2}, Sunwoo Lee\textsuperscript{4}*}
\\ \\
\textsuperscript{1}University of Southern California, United States \\
\textsuperscript{2}Neural Superintelligence Lab, MODULABS, Republic of Korea \\
\textsuperscript{3}Seoul National University, Republic of Korea \\
\textsuperscript{4}Inha University, Republic of Korea \\
\href{https://github.com/daniel-eai/LoRP-Locality-Aware-Redundancy-Pruning}
{\faGithub\ GitHub}
}

\begin{document}
\maketitle
\begin{abstract}

Large language models are known to contain representational redundancy across network depth, making depth pruning an effective approach for improving inference efficiency. Existing one-shot pruning methods rely on local layer importance or fixed redundancy assumptions across architectures. We propose Locality-Aware Redundancy Pruning (LoRP), a training-free one-shot depth pruning framework guided by representation locality. We show that inter-layer redundancy can be either localized or globally distributed depending on the LLM architecture. To characterize this phenomenon, we introduce Representation Locality Score (RLS), derived from global inter-layer hidden-state similarity. Using a small calibration set, LoRP computes pairwise layer similarity, clusters layers by representational similarity, and allocates pruning according to residual intra-cluster redundancy. Experiments across diverse LLM families show improvements in both perplexity and downstream task accuracy.

\begingroup
\renewcommand\thefootnote{}
\footnote{$^{\dag}$These authors contributed equally. $^{*}$Corresponding author}
\addtocounter{footnote}{-1}
\endgroup
\end{abstract}

\section{Introduction}
\label{sec:intro}
Large language models (LLMs) have achieved remarkable performance across a wide range of reasoning and language understanding tasks~\citep{brown2020gpt3, touvron2023llama, dubey2024llama3}. However, these advances come with substantial computational and memory costs due to the extreme depth and overparameterization of modern Transformer architectures~\citep{vaswani2017attention}. Consequently, improving inference efficiency has become an increasingly important research direction~\citep{zhu2023survey}. Among various compression techniques, layer-wise pruning is particularly attractive because removing entire Transformer blocks directly reduces both inference latency and memory usage~\citep{sleb, men2024shortgpt}.
Layer-wise pruning is also useful for compressing a fixed pretrained checkpoint. It provides flexible intermediate model sizes without retraining when a suitable smaller dense counterpart with the same architecture and training recipe is unavailable.

Recent studies have shown that many Transformer layers can be removed with relatively small performance degradation~\citep{men2024shortgpt, kim2024shortened}, suggesting significant redundancy across network depth. Existing one-shot depth pruning methods commonly rely on fixed pruning priors. Local-importance based methods~\citep{men2024shortgpt, kim2024shortened} estimate redundancy from isolated layer scores, while contiguous similarity based methods~\citep{chen2024llmstreamline, gromov2024deeper} assume that redundancy is concentrated within adjacent depth regions. However, our inter-layer similarity analysis (Sec.~\ref{sec:prob}) shows that redundancy can be either localized or globally distributed across depth depending on the architecture. This suggests that effective depth pruning should be guided by global, architecture-dependent representation geometry rather than by isolated layer scores or fixed contiguous regions.

To this end, we propose \textit{Locality-Aware Redundancy Pruning (LoRP)}, a training-free, one-shot depth pruning framework guided by the representation locality of each LLM architecture. Using a small calibration dataset, LoRP computes pairwise cosine similarity between Transformer layer activations and estimates representation locality using the Representation Locality Score (RLS), a metric derived from global inter-layer hidden-state similarity. Guided by the RLS, LoRP partitions the network into clusters of representationally similar layers~\citep{ng2001spectral, vonluxburg2007tutorial} and allocates the pruning budget through a two-stage procedure: it first distributes removals across clusters, then assigns the remaining budget according to residual redundancy. This produces pruning patterns that reflect the redundancy structure of each architecture.

Experiments across diverse LLM families show that LoRP consistently improves both perplexity and downstream reasoning performance under no-recovery pruning settings. More broadly, our findings suggest that redundancy in large language models is architecture-dependent and emerges as a structured property of representation geometry across network depth.

\noindent \textbf{Our contributions} are summarized as follows:
\begin{itemize}
    \item We show that representation locality differs substantially across LLM families, with some architectures exhibiting localized redundancy while others exhibit more globally distributed representational similarity across depth.
    
    \item We introduce the Representation Locality Score (RLS), a simple metric that quantifies representation locality from inter-layer hidden-state similarity.
    
    \item We propose Locality-Aware Redundancy Pruning (LoRP), a training-free, one-shot depth pruning framework that clusters layers by representational similarity and allocates pruning according to each architecture's redundancy structure.
    
    \item We show that LoRP consistently improves both perplexity and downstream reasoning across diverse LLM families without any post-pruning recovery.
\end{itemize}

\section{Related works}
\label{sec:rw}

\textbf{Pruning paradigms.}
LLM pruning methods are broadly divided into unstructured and structured pruning. Unstructured pruning removes individual weights~\citep{wanda, sparsegpt, spase, one-shot}, but often requires specialized kernels for practical speedups. Structured pruning instead removes architectural components such as attention heads, hidden dimensions, or Transformer blocks, producing smaller dense models that remain efficient on standard hardware. Among these approaches, depth pruning is particularly attractive because removing Transformer blocks directly reduces inference latency and memory usage~\citep{sleb}.

\noindent \textbf{Structured and one-shot depth pruning.}
Existing structured pruning methods remove architectural components using importance criteria derived from weights~\citep{ma2023llmpruner}, activations~\citep{an2024flap, ashkboos2024slicegpt}, or gradients~\citep{xia2024sheared}. Width pruning often requires additional fine-tuning or distillation~\citep{muralidharan2024compact}, whereas depth pruning preserves the original Transformer structure and tensor shapes, making it suitable for training-free compression.

Recent one-shot depth pruning methods primarily rely on local importance estimation or contiguous redundancy assumptions. ShortGPT~\citep{men2024shortgpt} uses Block Influence scores, while Shortened LLaMA~\citep{kim2024shortened} and Gromov et al.~\citep{gromov2024deeper} evaluate isolated layer removal effects. LLM-Streamline~\citep{chen2024llmstreamline} removes contiguous depth regions using boundary activation similarity. Recent work also suggests that redundancy depends on the calibration objective rather than a universal layer ranking~\citep{layerpruning}.

These methods generally assume similar redundancy structures across architectures. However, our study shows that representation locality differs substantially across LLM families, motivating architecture-dependent pruning behavior.

\section{Preliminary}
\label{sec:prob}
\begin{figure*}[t]
\centering
\includegraphics[width=1.0\linewidth]{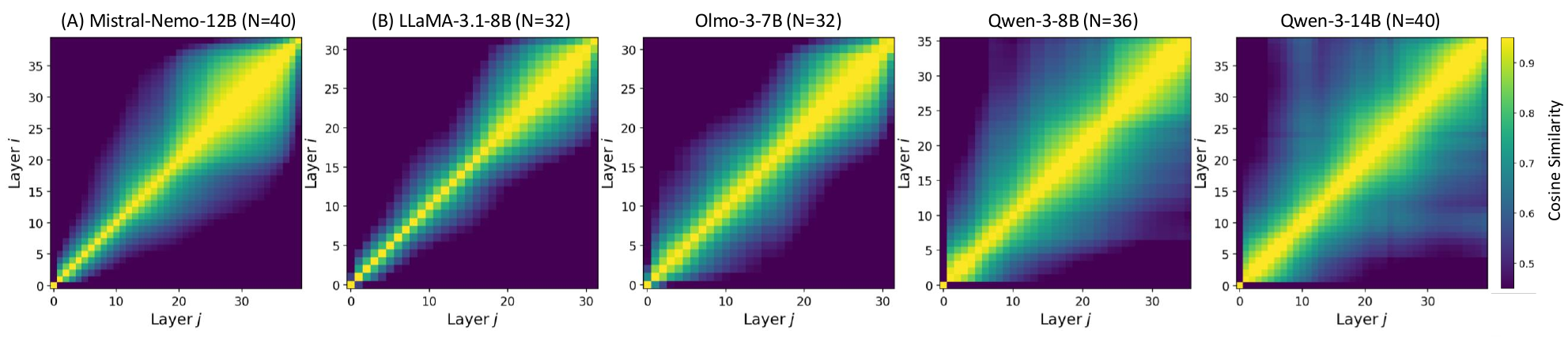}
\caption{
Pairwise inter-layer hidden-state cosine similarity matrices $S$ for five pre-trained LLMs. 
For each model, input hidden states from every Transformer block are collected using forward hooks over 128 calibration sequences from C4 with sequence length 2048. 
Each entry $S_{ij}$ denotes the mean per-token cosine similarity between the input representations of layers $i$ and $j$.
}
\label{fig:heatmap}
\end{figure*}

\subsection{Problem Setup}

Let
$
f_{\theta}
=
f_N \circ f_{N-1} \circ \cdots \circ f_1
$
denote a pretrained decoder-only Transformer consisting of $N$ sequential Transformer blocks, where $f_l$ represents the $l$-th block and $\theta$ denotes the full parameter set.

Given a target pruning budget $k \ll N$, the objective of depth pruning is to identify a subset of removable layers
\begin{equation}
    \mathcal{P}
    \subseteq
    \{1,\dots,N\},
    \qquad
    |\mathcal{P}| = k,
\end{equation}
such that the resulting compressed model
$
f_{\theta \setminus \mathcal{P}}
$
preserves the functional behavior of the original network while reducing inference-time computation and memory cost.

Formally, the ideal pruning objective can be written as
\begin{equation}
    \mathcal{P}^{*}
    =
    \arg\min_{\substack{\mathcal{P}\subseteq\{1,\dots,N\} \\ |\mathcal{P}|=k}}
    \mathcal{L}
    \left(
    f_{\theta},
    f_{\theta \setminus \mathcal{P}}
    \right),
\end{equation}
where
$
\mathcal{L}(\cdot,\cdot)
$
measures the functional discrepancy between the original and pruned models.

Since directly solving this objective is computationally intractable, practical training-free pruning methods instead rely on proxy redundancy measures to estimate removable layers.

\begin{figure}[h]
\centering
\includegraphics[width=0.8\linewidth]{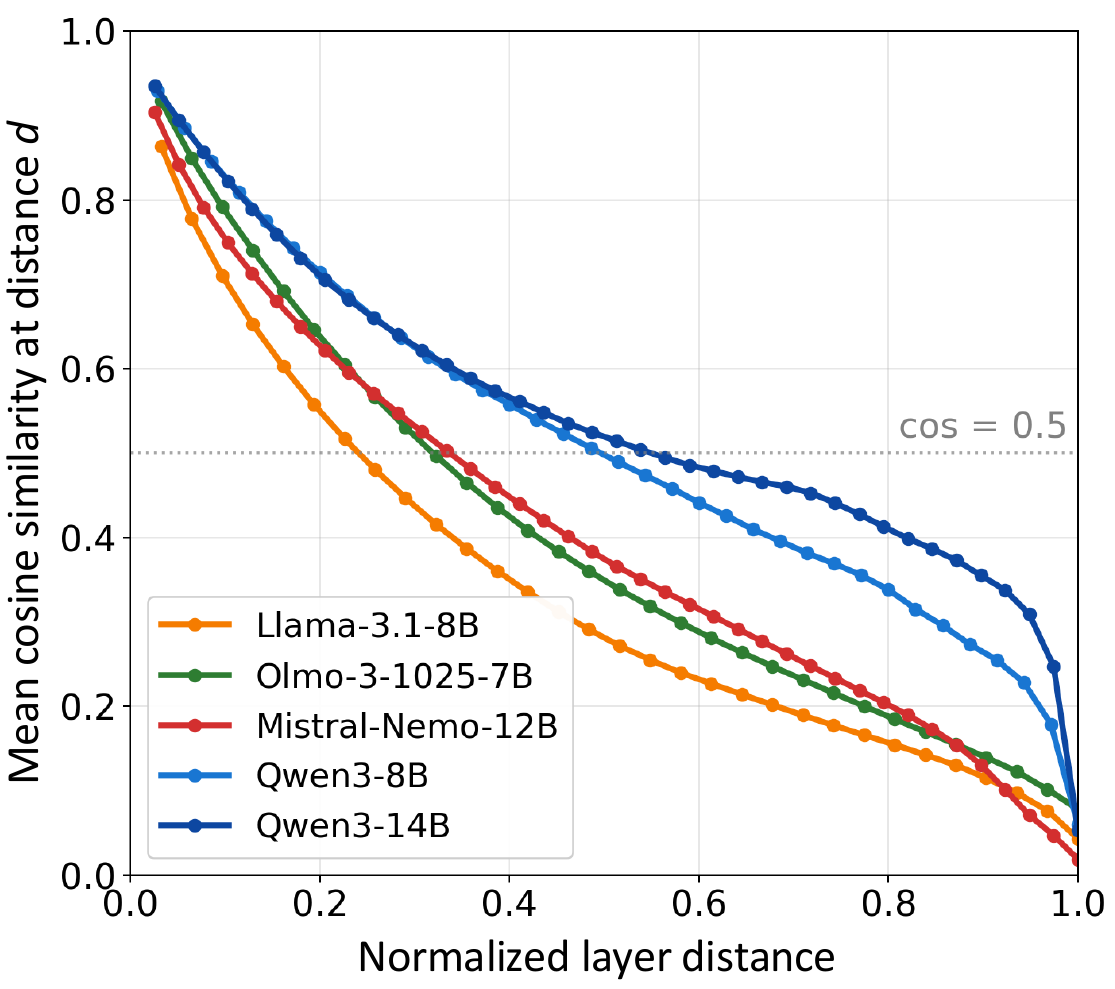}
\caption{
Mean inter-layer hidden-state cosine similarity as a function of normalized layer distance. 
For each model, $S_{ij}$ denotes the mean per-token cosine similarity between the input hidden states of layers $i$ and $j$, computed from 128 calibration sequences from C4. 
The x-axis represents normalized layer distance, and the y-axis shows the mean cosine similarity between all layer pairs separated by the corresponding distance.
}
\label{fig:cosime}
\end{figure}

\subsection{Motivational Observation}

Existing training-free depth-pruning methods typically assume similar redundancy structures across architectures, relying either on isolated layer importance or contiguous redundancy assumptions. However, we observe that representation locality differs substantially across LLM families. As shown in Figure~\ref{fig:heatmap} and Figure~\ref{fig:cosime}, Qwen models exhibit more globally distributed inter-layer similarity, whereas Llama, OLMo, and Mistral show more localized redundancy across depth.

These observations suggest that effective pruning behavior should depend on the representation locality of each architecture rather than a fixed redundancy assumption. Motivated by this, we model redundancy through global inter-layer representation similarity and design a pruning framework that adaptively allocates pruning according to the locality structure of each model. We next describe how this principle motivates our method.

\section{Method}
\label{sec:method}
\begin{figure*}[t]
\centering
\includegraphics[width=1.0\linewidth]{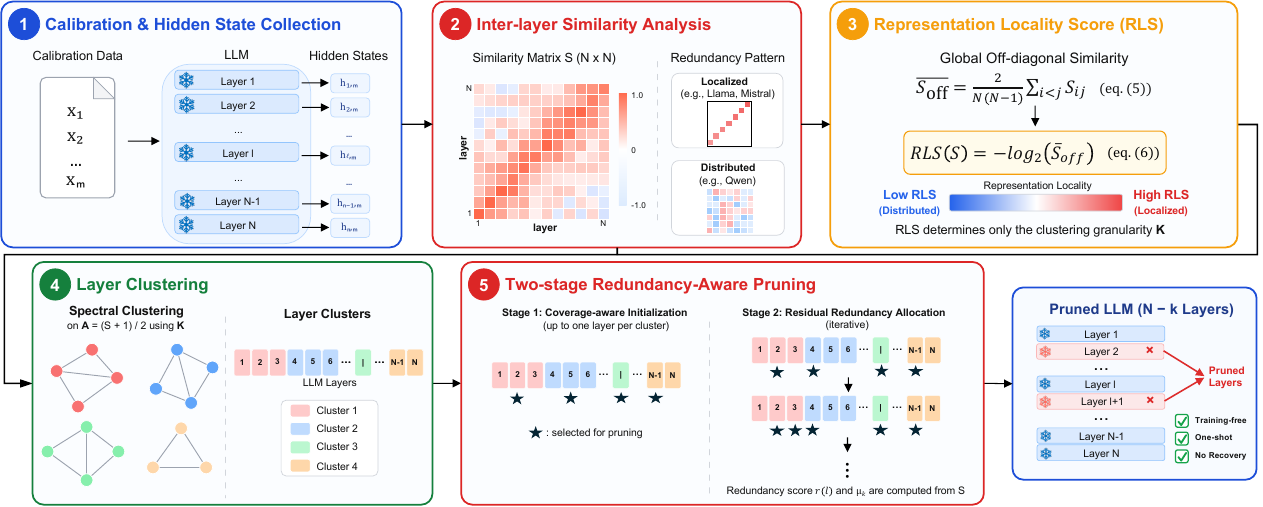}
\caption{
Overview of LoRP. 
\textbf{(1)} Input hidden states are collected from every Transformer block using a small calibration set. 
\textbf{(2)} The full inter-layer similarity matrix $S$ captures pairwise representation structure across depth. 
\textbf{(3)} RLS summarizes global representation locality and selects the clustering granularity $K$. 
\textbf{(4)} Spectral clustering partitions layers using the full affinity matrix $A=(S+1)/2$ and the selected $K$. 
\textbf{(5)} Stage 1 uses layer-level redundancy computed from $S$ to initialize the pruning set with up to one layer per cluster. Stage 2 allocates the remaining budget using residual cluster- and layer-level redundancy computed from $S$.
}
\label{fig:method}
\end{figure*}

We now introduce Locality-Aware Redundancy Pruning (LoRP), a training-free one-shot depth pruning framework motivated by the architecture-dependent representation locality discussed in Section~3.2. 
Figure~\ref{fig:method} illustrates the overall LoRP workflow.
LoRP first measures global inter-layer similarity, then partitions layers into representational clusters, and finally allocates the pruning budget through a two-stage redundancy-aware procedure.

\subsection{Inter-Layer Similarity and Locality}
\label{subsec:similarity}

\paragraph{Activation Similarity Measurement.}
Let
$
\mathcal{D}
=
\{x_m\}_{m=1}^{M}
$
denote an unlabeled calibration corpus. For every Transformer block $f_l$, we attach a forward hook and collect the input hidden-state activations. For token position $t$ in sample $x_m$, let
$
\mathbf{h}_{l,m,t}
\in
\mathbb{R}^{d}
$
denote the hidden representation entering layer $l$.

We normalize each token representation as
\begin{equation}
    \hat{\mathbf{h}}_{l,m,t}
    =
    \frac{
    \mathbf{h}_{l,m,t}
    }{
    \|\mathbf{h}_{l,m,t}\|_2 + \epsilon
    },
\end{equation}
where $\epsilon > 0$ is a small numerical constant.

We then compute the pairwise layer similarity matrix
$\mathbf{S}\in\mathbb{R}^{N\times N}$
by averaging token-wise cosine similarity across the calibration corpus:
\begin{equation}
\label{eq:similarity}
    S_{ij}
    =
    \frac{
    1
    }{
    \sum_{m=1}^{M} T_m
    }
    \sum_{m=1}^{M}
    \sum_{t=1}^{T_m}
    \hat{\mathbf{h}}_{i,m,t}^{\top}
    \hat{\mathbf{h}}_{j,m,t},
\end{equation}
where $T_m$ denotes the sequence length of sample $x_m$.


A larger value of $S_{ij}$ indicates that layers $i$ and $j$ process highly similar residual-stream representations over the calibration distribution. High similarity does not necessarily mean that two layers perform the same function. Instead, it indicates overlap in their hidden representations. We use this overlap as a signal for redundancy-aware pruning. This similarity score will be used to characterize representation locality and to partition layers into representational clusters in the following subsections.

\paragraph{Representation Locality Score (RLS).}
To characterize architecture-dependent representation locality, we define the Representation Locality Score (RLS), which measures the global average inter-layer similarity.

We first define the global off-diagonal similarity mean:
\begin{equation}
    \bar{S}_{\mathrm{off}}
    =
    \frac{
    2
    }{
    N(N-1)
    }
    \sum_{i<j}
    S_{ij}.
\end{equation}

We then define RLS as
\begin{equation}
\label{eq:rls}
    \mathrm{RLS}(\mathbf{S})
    =
    -\log_2
    \bar{S}_{\mathrm{off}}.
\end{equation}

A larger RLS indicates that inter-layer similarity decays more rapidly across depth, suggesting that redundancy is concentrated within localized regions of the network. Conversely, a smaller RLS reflects stronger global inter-layer similarity and more broadly distributed redundancy across depth.

\begin{table}[h]
\centering
\label{tab:rls_simple}
\small
\begin{tabular}{l c c}
\toprule
Model & \# Layers & RLS \\
\midrule
Llama-3.1-8B      & 32 & $\mathbf{1.149}$ \\
Olmo-3-7B         & 32 & $\mathbf{0.941}$ \\
Mistral-Nemo-12B  & 40 & $\mathbf{0.926}$ \\
Qwen3-8B          & 36 & $\mathbf{0.685}$ \\
Qwen3-14B         & 40 & $\mathbf{0.644}$ \\
\bottomrule
\end{tabular}
\caption{
Representation Locality Score (RLS) across different LLM families.
Higher RLS indicates more localized representation similarity across depth.
}
\end{table}


We use the global off-diagonal mean rather than only near-diagonal similarity so that RLS measures representation locality without assuming a specific redundancy structure across depth. Restricting the measurement to near-diagonal entries may bias the metric toward more localized similarity patterns, whereas aggregating across all layer pairs provides a broader measure of global inter-layer similarity. Importantly, RLS is intended as a scalar summary rather than a complete descriptor of the similarity matrix topology. In LoRP, RLS is used only to select the clustering granularity \(K\), while the actual layer partition and pruning scores are computed from the full similarity matrix \(S\). Consequently, two models with identical RLS values but different similarity structures may still produce different clusters and pruning decisions.

\subsection{Layer Clustering}
\label{subsec:clustering}


Using the inter-layer similarity matrix S and the Representation Locality Score (RLS) defined in Section 4.1, we partition the layers into representational clusters. The similarity matrix S provides the pairwise structure for grouping representationally similar layers, while RLS controls the partition granularity. Since cosine similarity may contain negative values, we first convert S into a non-negative affinity matrix:
\begin{equation}
    \mathbf{A}
    =
    \frac{\mathbf{S}+1}{2}.
\end{equation}

We then apply spectral clustering with precomputed affinity to partition the layers into $K$ disjoint representational clusters
$\mathcal{C}=\{C_1,C_2,\dots,C_K\}$.

Clusters are re-indexed according to first-occurrence depth so that earlier clusters receive smaller indices. The partition granularity is guided by RLS: lower RLS encourages finer-grained partitioning that spreads pruning decisions across depth, whereas higher RLS uses coarser partitioning concentrated within redundant regions. Spectral clustering further allows the partition to reflect non-adjacent representational similarity without assuming layer contiguity.

\subsection{Redundancy-Aware Pruning}
\label{subsec:pruning}

\paragraph{Intra-Cluster Redundancy Estimation.}
Given the representational partition $\mathcal{C}$, we estimate the redundancy of each layer relative to the remaining members of its cluster.

For a layer $l\in C_k$, we define the intra-cluster redundancy score as
\begin{equation}
\label{eq:redundancy}
    r(l)
    =
    \frac{
    1
    }{
    |C_k|-1
    }
    \sum_{\substack{m\in C_k \\ m\neq l}}
    S_{lm}.
\end{equation}

Layers with larger redundancy scores exhibit stronger representational overlap with other members of the same cluster and are therefore more likely to be removable with limited functional disruption. Within each cluster, layers are sorted in descending order of redundancy:
$
    r(l_k^{(1)})
    \geq
    r(l_k^{(2)})
    \geq
    \cdots
    $ where $
    l_k^{(j)} \in C_k.
$ Following prior observations in depth pruning, we exclude the boundary layers $\mathcal{B}=\{1,N\}$ from the pruning candidate set.

\paragraph{Two-Stage Allocation.}
LoRP performs pruning using a two-stage redundancy allocation strategy that balances representational coverage with residual redundancy estimation.

\noindent
\textit{Stage 1: Coverage-aware initialization.}
Rather than concentrating pruning decisions within a single highly redundant region, we first encourage representational coverage across clusters. For each cluster, we identify its most redundant eligible layer:
\begin{equation}
    l_k^{*}
    =
    \arg\max_{l\in C_k\setminus\mathcal{B}}
    r(l),
\end{equation}
and construct an initial candidate pool $
    \mathcal{P}_0
    =
    \bigcup_{k=1}^{K}
    \{l_k^{*}\}.
$ If the pruning budget is smaller than the number of clusters, we retain only the top-ranked candidates according to intra-cluster redundancy. 

\noindent \textit{Stage 2: Residual redundancy allocation.}
Starting from the initialized pruning set $\mathcal{P}\leftarrow\mathcal{P}_0$, we iteratively allocate the remaining pruning budget according to residual cluster redundancy.

For cluster $C_k$, let
$ \widetilde{C}_k
    =
    C_k
    \setminus
    \mathcal{P}
    \setminus
    \mathcal{B}
$ denote the remaining eligible members. We define the residual redundancy of cluster $C_k$ as the mean pairwise similarity among its remaining eligible layers:
\begin{equation}
\label{eq:cluster_energy}
    \mu_k
    =
    \frac{
    2
    }{
    |\widetilde{C}_k|
    (|\widetilde{C}_k|-1)
    }
    \sum_{\substack{i,j\in\widetilde{C}_k \\ i<j}}
    S_{ij}.
\end{equation}
if $|\widetilde{C}_k|<2$, we set $\mu_k=-\infty$.

At each iteration, we select the cluster with the largest residual redundancy $k^{*}
=
\arg\max_k
\mu_k,$
and prune the most redundant remaining layer from that cluster:
\begin{equation}
l^{*}
=
\arg\max_{l\in\widetilde{C}_{k^{*}}}
r(l).
\end{equation}

The pruning set is updated as $\mathcal{P}
\leftarrow
\mathcal{P}
\cup
\{l^{*}\},$ until the target pruning budget is reached.

The two-stage design avoids pruning collapse into a single cluster while adaptively updating redundancy after each pruning step. As a result, architectures with localized redundancy naturally continue pruning within the same cluster, whereas architectures with distributed redundancy tend to spread pruning decisions across multiple clusters.


\section{Experimental Results}
\label{sec:result}
\label{sec:results}

\begin{table*}[t]
\centering
\scriptsize
\setlength{\tabcolsep}{7pt}
\begin{adjustbox}{width=0.9\textwidth}
\begin{tabular}{l rrrr rrrr}
\toprule
 & \multicolumn{4}{c}{$L_p / L_t$} & \multicolumn{4}{c}{$L_p / L_t$} \\
\cmidrule(lr){2-5}\cmidrule(lr){6-9}
Method & Wiki & C4 & PTB & \textbf{AVG} \textcolor{red}{$\downarrow$}\ & Wiki & C4 & PTB & \textbf{AVG} \textcolor{red}{$\downarrow$}\ \\
\midrule
\textbf{LLaMA-3.1-8B}   & \multicolumn{4}{c}{7/32} & \multicolumn{4}{c}{9/32}\\
\midrule
Dense          & 6.24 & 8.68 & 10.58 & 8.50 & 6.24 & 8.68 & 10.58 & 8.50 \\
ShortGPT       & 63.42 & 69.64 & 68.58 & 67.21 & $\geq$2000 & $\geq$2000 & $\geq$2000 & $\geq$2000 \\
LaCo           & 77.82 & 66.82 & 88.05 & 77.56 & 1960.11 & 253.83 & 602.81 & 938.92 \\
LLM-Streamline & $\geq$2000 & 1173.53 & $\geq$2000 & $\geq$2000 & 210.85 & 155.23 & 182.96 & 183.01 \\
\jygl LoRP (ours)    & \jygl 27.88 & \jygl 35.41 & \jygl 62.56 & \jygl \textbf{41.95} & \jygl 155.11 & \jygl 124.82 & \jygl 175.79 & \jygl \textbf{151.91} \\
\midrule
\textbf{OLMo-3-7B}  & \multicolumn{4}{c}{7/32} & \multicolumn{4}{c}{9/32}\\
\midrule
Dense          & 9.92 & 15.70 & 17.40 & 14.34 & 9.92 & 15.70 & 17.40 & 14.34 \\
ShortGPT       & 24.25 & 30.24 & 35.55 & 30.02 & 41.21 & 45.01 & 57.87 & 48.03 \\
LaCo           & 21.84 & 27.63 & 39.81 & 29.76 & 46.23 & 43.02 & 81.58 & 56.94 \\
LLM-Streamline & 24.25 & 30.24 & 35.55 & 30.02 & 41.21 & 45.01 & 57.87 & 48.03 \\
\jygl LoRP (ours)    & \jygl 18.66 & \jygl 25.85 & \jygl 29.94 & \jygl \textbf{24.82} & \jygl 29.27 & \jygl 35.39 & \jygl 57.40 & \jygl \textbf{40.69} \\
\midrule
\textbf{Qwen3-8B}  & \multicolumn{4}{c}{7/36} & \multicolumn{4}{c}{9/36}\\
\midrule
Dense          & 9.71 & 14.45 & 16.90 & 13.69 & 9.71 & 14.45 & 16.90 & 13.69 \\
ShortGPT       & 230.98 & 88.73 & 468.29 & 262.67 & 1074.21 & 313.33 & 1776.87 & 1054.80 \\
LaCo           & 270.75 & 83.03 & 215.04 & 189.61 & 1595.08 & 397.32 & 831.80 & 941.40 \\
LLM-Streamline & 217.45 & 83.00 & 386.89 & 229.11 & 1074.21 & 313.33 & 1776.87 & 1054.80 \\
\jygl LoRP (ours)    & \jygl 22.18 & \jygl 27.38 & \jygl 34.22 & \jygl \textbf{27.92} & \jygl 213.19 & \jygl 66.99 & \jygl 174.73 & \jygl \textbf{151.64} \\
\midrule
\textbf{Qwen3-14B} & \multicolumn{4}{c}{11/40} & \multicolumn{4}{c}{13/40}\\
\midrule
Dense          & 8.64 & 13.00 & 14.79 & 12.14 & 8.64 & 13.00 & 14.79 & 12.14 \\
ShortGPT       & 122.11 & 112.97 & 594.40 & 276.50 & 1875.14 & 391.47 & $\geq$2000 & $\geq$2000 \\
LaCo           & 823.83 & 688.82 & 1561.39 & 1024.68 & $\geq$2000 & 1971.89 & $\geq$2000 & $\geq$2000 \\
LLM-Streamline & $\geq$2000 & 1612.04 & $\geq$2000 & $\geq$2000 & $\geq$2000 & $\geq$2000 & $\geq$2000 & $\geq$2000 \\
\jygl LoRP (ours)    & \jygl 27.11 & \jygl 33.47 & \jygl 51.73 & \jygl \textbf{37.44} & \jygl 61.10 & \jygl 68.71 & \jygl 100.28 & \jygl \textbf{76.69} \\
\midrule
\textbf{Mistral-Nemo-12B} & \multicolumn{4}{c}{11/40} & \multicolumn{4}{c}{13/40}\\
\midrule
Dense          & 5.75 & 9.09 & 26.70 & 13.85 & 5.75 & 9.09 & 26.70 & 13.85 \\
ShortGPT       & 161.78 & 62.41 & 849.59 & 357.93 & 284.66 & 160.96 & 771.43 & 405.68 \\
LaCo           & 66.70 & 63.83 & 199.77 & 110.10 & 110.71 & 90.23 & 468.07 & 223.01 \\
LLM-Streamline & 161.78 & 62.41 & 849.59 & 357.93 & 121.37 & 66.81 & 731.84 & 306.67 \\
\jygl LoRP (ours)    & \jygl 22.86 & \jygl 26.49 & \jygl 103.24 & \jygl \textbf{50.86} & \jygl 111.13 & \jygl 97.22 & \jygl 411.07 & \jygl \textbf{206.47} \\
\bottomrule
\end{tabular}
\end{adjustbox}
\caption{
Perplexity ($\downarrow$) on WikiText-2 (Wiki), C4, and Penn Treebank (PTB) after training-free depth pruning without recovery.
Avg.\ denotes the arithmetic mean across the three datasets, and
$L_p/L_t$ denotes the number of pruned layers over the total number of Transformer blocks.
Lower values indicate better preservation of language modeling behavior after pruning.
Values $\geq 2000$ are capped for readability.
}
\label{tab:ppl}
\end{table*}


\begin{table*}[h]
\centering
\tiny
\vspace{2pt}
\begin{adjustbox}{width=0.9\textwidth}
\begin{tabular}{l l l c c c c c c c c c r}
\toprule
\textbf{Model} & \textbf{$L_p/L_t$} & \textbf{Method} & \textbf{ARC-E} & \textbf{ARC-C} & \textbf{HellaS} & \textbf{WinoG} & \textbf{BoolQ} & \textbf{OBQA} & \textbf{RTE} & \textbf{CoPa} & \textbf{Race} & \textbf{AVG} \textcolor{red}{$\uparrow$} \\
\midrule
\multirow{9}{*}{\begin{sideways}LLaMA-3.1-8B\end{sideways}}
& 0/32 & Dense          & 81.31 & 53.50 & 78.90 & 73.72 & 82.02 & 44.80 & 69.68 & 87.00 & 39.23 & 67.79 \\
\cmidrule{2-13}
& 7/32 & LLM-Streamline & 44.15 & 33.02 & 33.40 & 56.83 & 38.20 & 32.60 & 58.12 & 61.00 & 26.03 & 42.59 \\
& 7/32 & ShortGPT       & 58.29 & 42.15 & 64.93 & 68.27 & 62.02 & 34.60 & 69.31 & 80.00 & 34.35 & 57.10 \\
& 7/32 & LaCo           & 57.58 & 40.53 & 64.86 & 69.06 & 68.04 & 36.00 & 66.79 & 76.00 & 33.78 & 56.96 \\
& 7/32 & \jygl LoRP (Ours) & \jygl 62.37 & \jygl 42.66 & \jygl 63.63 & \jygl 65.98 & \jygl 66.21 & \jygl 39.20 & \jygl 59.21 & \jygl 80.00 & \jygl 35.60 & \jygl \textbf{57.21} \\
\cmidrule{2-13}
& 9/32 & LLM-Streamline & 49.83 & 37.80 & 56.03 & 65.98 & 58.87 & 32.80 & 67.15 & 69.00 & 30.24 & 51.97 \\
& 9/32 & ShortGPT       & 39.18 & 30.72 & 29.06 & 53.91 & 38.26 & 29.40 & 67.15 & 59.00 & 25.07 & 41.30 \\
& 9/32 & LaCo           & 49.07 & 36.09 & 51.99 & 63.30 & 65.05 & 35.20 & 68.23 & 71.00 & 31.77 & \textbf{52.41} \\
& 9/32 & \jygl LoRP (Ours) & \jygl 52.82 & \jygl 38.23 & \jygl 55.93 & \jygl 63.14 & \jygl 56.48 & \jygl 32.20 & \jygl 62.82 & \jygl 75.00 & \jygl 31.96 & \jygl 52.06 \\
\midrule
\multirow{9}{*}{\begin{sideways}OLMo-3-7B\end{sideways}}
& 0/32 & Dense          & 78.07 & 52.56 & 74.91 & 69.46 & 81.04 & 41.80 & 72.20 & 87.00 & 39.04 & 66.23 \\
\cmidrule{2-13}
& 7/32 & LLM-Streamline & 65.74 & 42.06 & 61.60 & 67.17 & 73.36 & 36.00 & 70.04 & 77.00 & 35.60 & 58.73 \\
& 7/32 & ShortGPT       & 65.74 & 42.06 & 61.60 & 67.17 & 73.36 & 36.00 & 70.04 & 77.00 & 35.60 & 58.73 \\
& 7/32 & LaCo           & 67.09 & 39.93 & 62.84 & 66.93 & 70.12 & 35.60 & 68.23 & 75.00 & 36.46 & 58.02 \\
& 7/32 & \jygl LoRP (Ours) & \jygl 74.33 & \jygl 45.31 & \jygl 63.67 & \jygl 64.40 & \jygl 70.40 & \jygl 33.80 & \jygl 70.76 & \jygl 84.00 & \jygl 36.75 & \jygl \textbf{60.38} \\
\cmidrule{2-13}
& 9/32 & LLM-Streamline & 55.60 & 35.67 & 53.85 & 65.04 & 66.15 & 31.40 & 63.54 & 67.00 & 32.54 & 52.31 \\
& 9/32 & ShortGPT       & 55.60 & 35.67 & 53.85 & 65.04 & 66.15 & 31.40 & 63.54 & 67.00 & 32.54 & 52.31 \\
& 9/32 & LaCo           & 57.49 & 34.04 & 55.79 & 63.46 & 70.31 & 34.00 & 63.54 & 73.00 & 33.97 & 53.95 \\
& 9/32 & \jygl LoRP (Ours) & \jygl 68.14 & \jygl 41.38 & \jygl 58.32 & \jygl 64.72 & \jygl 72.60 & \jygl 33.20 & \jygl 74.37 & \jygl 74.00 & \jygl 37.03 & \jygl \textbf{58.20} \\
\midrule
\multirow{9}{*}{\begin{sideways}Mistral-Nemo-12B\end{sideways}}
& 0/40  & Dense          & 81.61 & 58.02 & 82.79 & 73.48 & 85.26 & 47.20 & 65.34 & 91.00 & 41.82 & 69.61 \\
\cmidrule{2-13}
& 11/40 & LLM-Streamline & 60.14 & 41.13 & 51.45 & 61.88 & 66.61 & 33.20 & 66.06 & 78.00 & 31.96 & 54.49 \\
& 11/40 & ShortGPT       & 60.14 & 41.13 & 51.45 & 61.88 & 66.61 & 33.20 & 66.06 & 78.00 & 31.96 & 54.49 \\
& 11/40 & LaCo           & 57.91 & 39.08 & 62.46 & 61.64 & 67.98 & 33.40 & 58.12 & 75.00 & 32.25 & 54.20 \\
& 11/40 & \jygl LoRP (Ours) & \jygl 58.21 & \jygl 38.05 & \jygl 62.10 & \jygl 69.06 & \jygl 65.20 & \jygl 32.20 & \jygl 58.48 & \jygl 76.00 & \jygl 36.46 & \jygl \textbf{55.08} \\
\cmidrule{2-13}
& 13/40 & LLM-Streamline & 50.67 & 35.49 & 49.90 & 66.93 & 69.79 & 31.80 & 55.96 & 64.00 & 31.39 & 50.66 \\
& 13/40 & ShortGPT       & 50.84 & 35.67 & 55.06 & 57.93 & 62.14 & 32.40 & 57.76 & 65.00 & 31.00 & 49.76 \\
& 13/40 & LaCo           & 53.87 & 35.24 & 57.70 & 60.14 & 63.91 & 32.80 & 60.65 & 70.00 & 31.48 & \textbf{51.76} \\
& 13/40 & \jygl LoRP (Ours) & \jygl 52.95 & \jygl 37.20 & \jygl 57.35 & \jygl 60.69 & \jygl 63.43 & \jygl 31.40 & \jygl 57.04 & \jygl 68.00 & \jygl 31.29 & \jygl 51.04 \\
\midrule
\multirow{9}{*}{\begin{sideways}Qwen3-8B\end{sideways}}
& 0/36 & Dense          & 80.93 & 56.48 & 74.95 & 67.56 & 86.61 & 41.40 & 78.34 & 85.00 & 41.24 & 68.06 \\
\cmidrule{2-13}
& 7/36 & LLM-Streamline & 51.98 & 37.97 & 55.24 & 60.22 & 82.60 & 33.40 & 78.34 & 55.00 & 31.87 & 54.07 \\
& 7/36 & ShortGPT       & 51.85 & 35.75 & 55.32 & 58.01 & 69.76 & 33.00 & 70.76 & 56.00 & 33.01 & 51.50 \\
& 7/36 & LaCo           & 56.57 & 38.74 & 54.01 & 59.59 & 81.38 & 35.20 & 80.51 & 63.00 & 33.40 & 55.82 \\
& 7/36 & \jygl LoRP (Ours) & \jygl 59.30 & \jygl 38.65 & \jygl 55.29 & \jygl 61.96 & \jygl 75.32 & \jygl 33.00 & \jygl 75.81 & \jygl 75.00 & \jygl 35.50 & \jygl \textbf{56.65} \\
\cmidrule{2-13}
& 9/36 & LLM-Streamline & 42.68 & 30.89 & 49.80 & 54.78 & 64.77 & 31.00 & 77.26 & 53.00 & 32.34 & 48.50 \\
& 9/36 & ShortGPT       & 42.68 & 30.89 & 49.80 & 54.78 & 64.77 & 31.00 & 77.26 & 53.00 & 32.34 & 48.50 \\
& 9/36 & LaCo           & 46.59 & 32.68 & 46.72 & 56.75 & 83.36 & 33.20 & 76.90 & 63.00 & 30.05 & 52.14 \\
& 9/36 & \jygl LoRP (Ours) & \jygl 51.01 & \jygl 33.02 & \jygl 48.75 & \jygl 57.46 & \jygl 78.75 & \jygl 32.20 & \jygl 72.92 & \jygl 64.00 & \jygl 33.78 & \jygl \textbf{52.43} \\
\midrule
\multirow{9}{*}{\begin{sideways}Qwen3-14B\end{sideways}}
& 0/40  & Dense          & 82.83 & 60.24 & 78.82 & 72.85 & 89.30 & 46.20 & 77.62 & 90.00 & 43.16 & 71.22 \\
\cmidrule{2-13}
& 11/40 & LLM-Streamline & 41.71 & 31.91 & 44.91 & 55.64 & 63.24 & 30.80 & 53.07 & 68.00 & 29.28 & 46.51 \\
& 11/40 & ShortGPT       & 55.30 & 37.71 & 48.85 & 58.25 & 64.16 & 30.20 & 53.43 & 72.00 & 34.83 & 50.53 \\
& 11/40 & LaCo           & 45.33 & 31.74 & 39.54 & 56.43 & 76.15 & 29.00 & 60.29 & 62.00 & 27.85 & 47.59 \\
& 11/40 & \jygl LoRP (Ours) & \jygl 64.02 & \jygl 38.23 & \jygl 54.50 & \jygl 60.06 & \jygl 70.03 & \jygl 34.40 & \jygl 54.51 & \jygl 64.00 & \jygl 35.79 & \jygl \textbf{52.84} \\
\cmidrule{2-13}
& 13/40 & LLM-Streamline & 35.73 & 25.09 & 38.45 & 52.41 & 63.79 & 28.20 & 53.07 & 58.00 & 27.75 & 42.50 \\
& 13/40 & ShortGPT       & 44.15 & 30.72 & 44.13 & 53.83 & 63.00 & 30.40 & 53.43 & 65.00 & 31.87 & 46.28 \\
& 13/40 & LaCo           & 39.81 & 30.38 & 38.16 & 53.51 & 59.91 & 28.60 & 57.04 & 66.00 & 25.84 & 44.36 \\
& 13/40 & \jygl LoRP (Ours) & \jygl 56.36 & \jygl 33.28 & \jygl 48.24 & \jygl 56.04 & \jygl 66.30 & \jygl 30.40 & \jygl 54.87 & \jygl 65.00 & \jygl 35.02 & \jygl \textbf{49.50} \\
\bottomrule
\end{tabular}
\end{adjustbox}
\caption{
Zero-shot accuracy (\%) on nine commonsense reasoning benchmarks after training-free depth pruning without recovery.
AVG denotes the mean accuracy across all tasks, and
$L_p/L_t$ denotes the number of pruned layers over the total number of Transformer blocks.
Higher values indicate better preservation of downstream reasoning performance after pruning.
}
\label{tab:commonsense_main}
\end{table*}

\subsection{Experimental Setup}
\label{sec:settings}

\noindent \textbf{Benchmarks.}
We evaluate LoRP on nine zero-shot reasoning benchmarks:
\texttt{ARC-Easy} and \texttt{ARC-Challenge}~\citep{clark2018arc},
\texttt{HellaSwag}~\citep{zellers2019hellaswag},
\texttt{WinoGrande}~\citep{sakaguchi2021winogrande},
\texttt{BoolQ}~\citep{clark2019boolq},
\texttt{OpenbookQA}~\citep{mihaylov2018openbookqa},
\texttt{RTE}~\citep{rte},
\texttt{COPA}~\citep{roemmele2011copa},
and \texttt{RACE}~\citep{lai2017race},
using the \texttt{lm-evaluation-harness}~\citep{gao2024lmeval}.
For perplexity evaluation, we report results on
\texttt{C4}~\citep{raffel2020c4}, \texttt{WikiText-2 (Wiki)}~\citep{merity2017wikitext} and \texttt{Penn Treebank (PTB)}~\citep{marcus1993ptb}
using sequence length 2048.

\noindent \textbf{Models.}
We evaluate LoRP on five open-source LLMs:
LLaMA-3.1-8B~\cite{dubey2024llama3}, OLMo-3-7B~\cite{olmo2025olmo2}, Mistral-Nemo-12B~\cite{mistral2024nemo},
Qwen3-8B~\cite{yang2025qwen3}, and Qwen3-14B~\cite{yang2025qwen3}.

\noindent \textbf{Baselines.}
We compare LoRP against three training-free depth pruning methods:
ShortGPT~\cite{men2024shortgpt}, LLM-Streamline~\cite{chen2024llmstreamline}, and LaCo~\cite{yang2024laco}.


\noindent \textbf{Calibration and implementation.}
The pairwise cosine similarity matrix $S$ is estimated using 128 sequences of length 2048 sampled from \texttt{C4}. Based on the ablation results in Section~\ref{ablation}, we use an RLS-guided clustering policy: $\mathrm{RLS} \geq 1.0$ uses $K{=}2$, $0.7 \leq \mathrm{RLS} < 1.0$ uses $K{=}3$, and $\mathrm{RLS} < 0.7$ uses $K{=}4$. We find that these RLS ranges consistently correspond to different effective clustering granularities across evaluated models. Consequently, we set $K{=}2$ for LLaMA-3.1-8B, $K{=}3$ for OLMo-3-7B and Mistral-Nemo-12B, and $K{=}4$ for Qwen3-8B,14B. A more detailed analysis of $K$ is provided in the ablation study in Section~\ref{ablation}. All experiments are conducted on a single NVIDIA A40 48GB GPU. We additionally report the computational cost reduction after depth pruning in Appendix~\ref{app:cost}.

\begin{figure*}[t]
\centering
\includegraphics[width=1\linewidth]{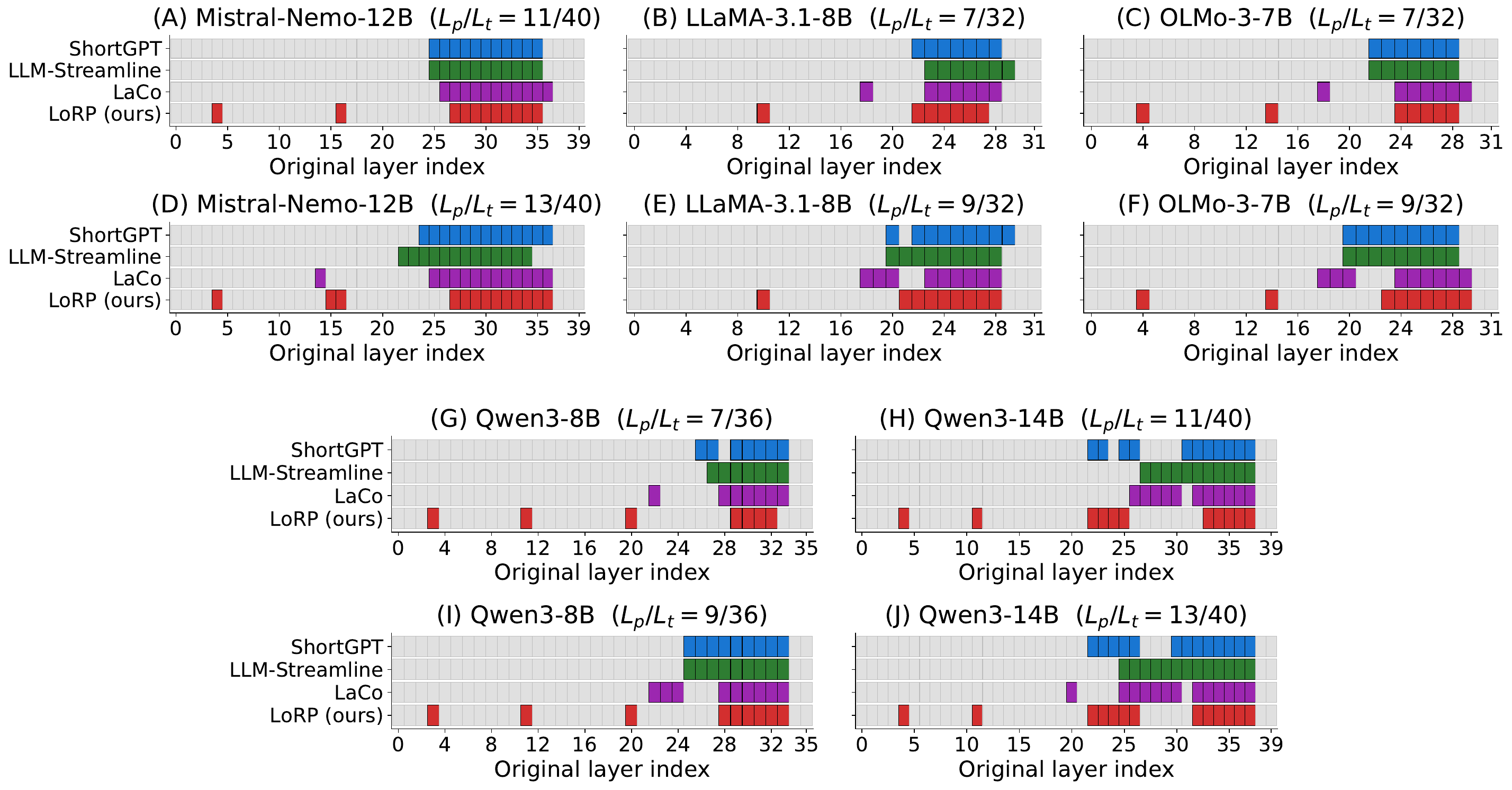}
\caption{
Visualization of pruned layer indices selected by different training-free depth pruning methods across LLM architectures and pruning budgets. Colored blocks indicate removed Transformer layers.
}
\label{fig:pattern}
\end{figure*}

\begin{table*}[t]
\centering
\footnotesize
\begin{adjustbox}{width=0.7\textwidth}
\begin{tabular}{l c c c l}
\toprule
Method & $L_p / L_t$ & PPL $\downarrow$ & DS $\uparrow$ & Pruned layers \\
\midrule
\multicolumn{1}{l}{\textbf{Llama-3.1-8B}} & 0/32 & 8.50  & 67.80 & -- \\
\hdashline
SLEB~\cite{sleb}            & 7/32 & 679.2 & 44.87 & [10--12, 19, 24--26] \\
Shortened LLaMA~\cite{kim2024shortened} & 7/32 & 18.60 & 49.44 & [8--12, 25,26] \\
LoRP (ours)     & 7/32 & 41.95 & 57.21 & [10, 22--27] \\
\hdashline
SLEB~\cite{sleb}            & 9/32 & 769.3 & 38.51 & [8--12, 19, 24--26] \\
Shortened LLaMA~\cite{kim2024shortened} & 9/32 & 34.41 & 46.77 & [6, 8--12, 25,26, 28] \\
LoRP (ours)     & 9/32 & 151.9 & 52.06 & [10, 21--28] \\
\midrule
\multicolumn{1}{l}{\textbf{Qwen3-8B}} & 0/36 & 13.69 & 68.06 & -- \\
\hdashline
SLEB~\cite{sleb}            & 7/36 & 78.99 & 48.49 & [2, 15,16, 20,21, 26, 32] \\
Shortened LLaMA & 7/36 & 21.06 & 54.87 & [2, 15--18, 20,21] \\
LoRP (ours)     & 7/36 & 27.92 & 56.65 & [3, 11, 20, 29--32] \\
\hdashline
SLEB~\cite{sleb}            & 9/36 & 330.6 & 42.49 & [2, 12, 14--16, 20,21, 26, 32] \\
Shortened LLaMA~\cite{kim2024shortened} & 9/36 & 26.48 & 51.73 & [2, 11, 15--21] \\
LoRP (ours)     & 9/36 & 151.6 & 52.43 & [3, 11, 20, 28--33] \\
\bottomrule
\end{tabular}
\end{adjustbox}
\caption{Iterative / per-layer pruning baselines vs.\ LoRP (ours) on Llama-3.1-8B and Qwen3-8B. PPL is averaged over C4 / WikiText-2 / PTB; DS is the mean accuracy on 9 zero-shot tasks.}
\label{tab:discussion-iterative-vs-ours}
\end{table*}


\subsection{Locality-Aware Pruning Improves Perplexity (PPL)}
Table~\ref{tab:ppl} evaluates perplexity after training-free depth pruning across different LLM architectures. Across most evaluated models and pruning budgets, LoRP achieves the lowest average perplexity among competing pruning methods. The improvement is particularly noticeable on Qwen models, where inter-layer similarity remains more broadly distributed across depth. In contrast, methods favoring localized or contiguous pruning often exhibit substantially larger perplexity degradation under aggressive pruning budgets. Overall, the results suggest that guiding pruning behavior according to architecture-dependent representation locality helps preserve language modeling performance after depth pruning.


\subsection{Locality-Aware Pruning Better Preserves Downstream Accuracy}
Table~\ref{tab:commonsense_main} evaluates downstream accuracy after training-free depth pruning across different LLM architectures. Compared to existing pruning methods, LoRP more preserves zero-shot reasoning performance across pruning budgets. The difference becomes particularly visible on architectures exhibiting more globally distributed redundancy patterns, such as the Qwen family, where localized or contiguous pruning strategies often lead to larger downstream degradation. Overall, the results suggest that locality-aware pruning allocation helps retain downstream generalization behavior after depth pruning.



\subsection{Architecture-Dependent Pruning Pattern Analysis}
Figure~\ref{fig:pattern} visualizes the layer indices selected by different pruning methods across models and pruning budgets. LoRP produces architecture-dependent pruning patterns that reflect the representation locality of each model. For models with more localized redundancy, such as LLaMA-3.1 and Mistral-Nemo, LoRP tends to remove layers from concentrated depth regions. In contrast, for Qwen models, where inter-layer similarity is more broadly distributed across depth, LoRP selects layers from multiple regions of the network. This behavior highlights the role of clustering in LoRP, which allocates pruning decisions according to both representational coverage and residual redundancy.


\subsection{Ablation Study}
\label{ablation}
\noindent\textbf{Iterative pruning vs LoRP.}
Table~\ref{tab:discussion-iterative-vs-ours} compares LoRP against iterative depth pruning baselines such as Shortened LLaMA and SLEB. Shortened LLaMA often achieves lower perplexity, reflecting stronger adaptation to the calibration objective through iterative pruning. In contrast, LoRP preserves higher downstream accuracy, particularly on Qwen3-8B, indicating better generalization beyond calibration-based layer selection. SLEB exhibits substantially worse perplexity and downstream performance than both Shortened LLaMA and LoRP across the evaluated settings, indicating that an iterative removal procedure alone does not guarantee robust pruning behavior. \\

\begin{figure}[h]
\centering
\includegraphics[width=1\linewidth]{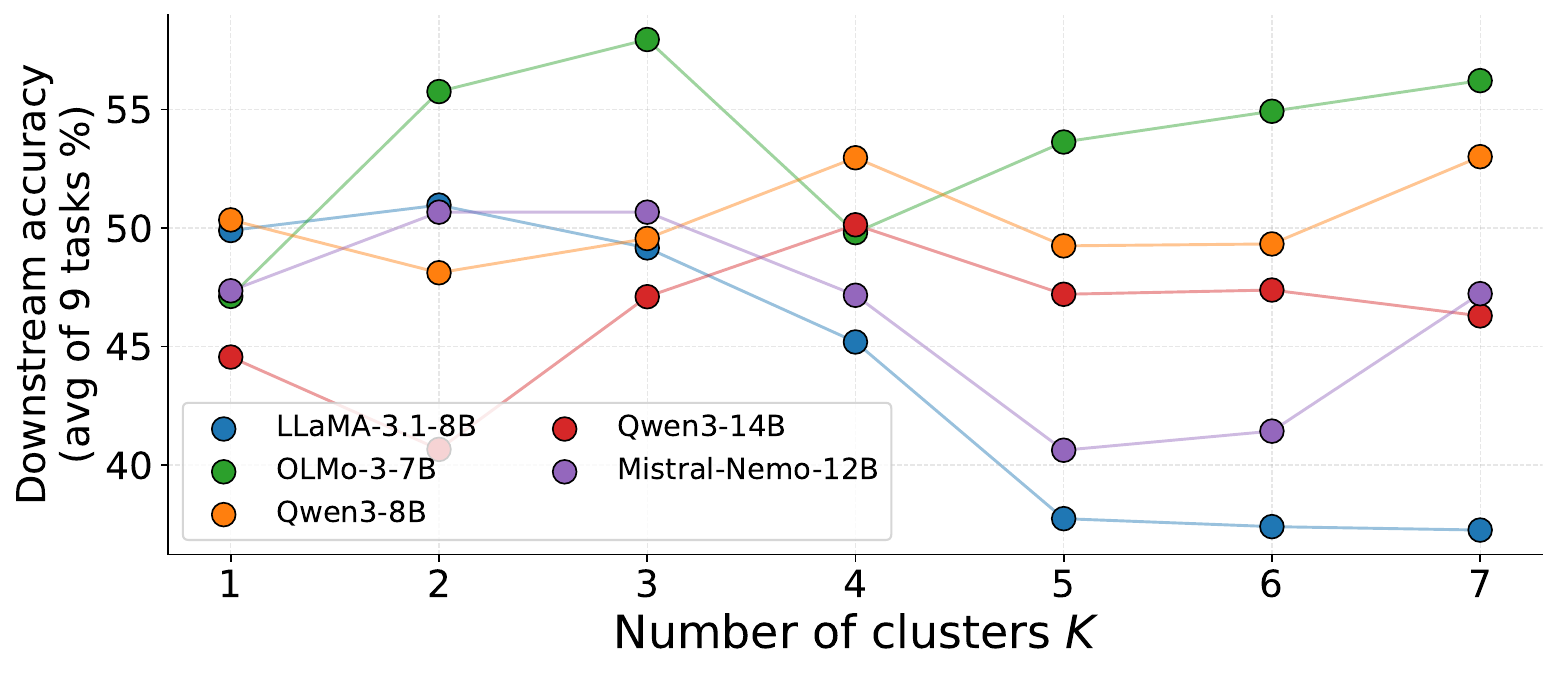}
\caption{Effect of the number of representational clusters $K$ on downstream accuracy across different LLM architectures under the same experimental setting as the main results.}
\label{fig:cluster}
\end{figure}

\noindent\textbf{Effect of the number of clusters.}
Figure~\ref{fig:cluster} analyzes the effect of the number of representational clusters $K$ on downstream accuracy. The best clustering granularity varies across architectures. Models with globally distributed redundancy, such as Qwen3, generally benefit from larger $K$. Models with more localized redundancy tend to perform well with smaller $K$. The $K$ selected by the fixed RLS-based policy achieves the best observed performance on LLaMA-3.1-8B, OLMo-3-7B, and Qwen3-14B. On Qwen3-8B and Mistral-Nemo-12B, its accuracy is within 0.10 percentage points of the best value in the sweep.

\noindent\textbf{Contribution of coverage-aware initialization.}
We isolate the contribution of Stage 1 by comparing full LoRP with a Stage-2-only variant. The latter begins residual redundancy allocation from an empty pruning set. As shown in Table~\ref{tab:stage-ablation}, full LoRP achieves lower C4 perplexity at every pruning budget on both LLaMA-3.1-8B and Qwen3-8B. This result shows that the initial coverage across representational clusters improves pruning robustness.

\begin{table}[t]
\centering
\small
\begin{tabular}{lccc}
\toprule
Model & $L_p$ & Full LoRP & Stage 2 only \\
\midrule
LLaMA-3.1-8B & 7  & \textbf{35.41}   & 56.28   \\
              & 9  & \textbf{124.82}  & 155.23  \\
              & 11 & \textbf{4280.59} & 6508.83 \\
\midrule
Qwen3-8B     & 7  & \textbf{27.38}  & 60.93  \\
              & 9  & \textbf{66.99}  & 96.91  \\
              & 11 & \textbf{124.38} & 183.91 \\
\bottomrule
\end{tabular}
\caption{C4 perplexity of full LoRP and the Stage-2-only variant. Lower values are better.}
\label{tab:stage-ablation}
\end{table}

\noindent\textbf{Calibration robustness.}
We evaluate the sensitivity of LoRP to the calibration configuration
by varying the sample count, sequence length, and corpus.
As shown in Table~\ref{tab:calibration-robustness},
LoRP selects identical layers across all evaluated settings
on both LLaMA-3.1-8B and Qwen3-8B.
These results indicate that LoRP is robust to calibration choices
within the evaluated range.

\begin{table}[t]
\centering
\small
\begin{tabular}{ll}
\toprule
Calibration setting & Evaluated values \tabularnewline
\midrule
Sample count
& 32, 64, 128, 256, 512 \tabularnewline
Sequence length
& 512, 1024, 2048 \tabularnewline
Corpus
& C4, WikiText-2 \tabularnewline
\midrule
Model & Selected layers \tabularnewline
\midrule
LLaMA-3.1-8B
& \{10, 22--27\} \tabularnewline
Qwen3-8B
& \{3, 11, 20, 29--32\} \tabularnewline
\bottomrule
\end{tabular}
\caption{Calibration robustness of LoRP with $L_p=7$.
Each model produces the same layer selection across all evaluated
calibration settings.}
\label{tab:calibration-robustness}
\end{table}



\section{Conclusion}
\label{sec:conclusion}
We proposed LoRP, a training-free one-shot depth pruning framework guided by architecture-dependent representation locality. We showed that redundancy structures differ substantially across LLMs, ranging from localized to globally distributed inter-layer similarity patterns. To capture this behavior, we introduced the Representation Locality Score (RLS) and used it to guide clustering-based redundancy-aware pruning. Across diverse LLM architectures, LoRP improved both perplexity and downstream accuracy under no-recovery pruning settings. Overall, our results suggest that representation locality is an important factor for effective depth pruning in large language models.

\section*{Limitations}
Although LoRP shows consistent improvements across diverse LLM families, our experiments remain limited to decoder-only architectures and a moderate range of model scales. The fixed RLS-to-\(K\) thresholds are empirically derived from the original evaluation models. While the policy selects the best-performing \(K\) under both pruning budgets on the held-out Gemma-2-9B model, broader validation across architectures, scales, and training recipes is still required.

The causal origins of localized and globally distributed redundancy also remain unclear. Architectural and training factors such as normalization, residual scaling, depth-to-width ratio, and the pretraining recipe may contribute to the observed patterns, but establishing causal relationships would require controlled pretraining experiments.

Across the tested calibration sample counts, sequence lengths, and C4/WikiText-2 corpora, LoRP produces identical layer selections on LLaMA-3.1-8B and Qwen3-8B. This demonstrates robustness within the evaluated range, but does not establish invariance across arbitrary domains or substantially different calibration distributions.

Our downstream evaluation primarily covers language understanding and commonsense reasoning tasks. We do not evaluate generative multi-step mathematical reasoning benchmarks such as GSM8K and MATH, and pruning behavior on these tasks remains an important direction for future evaluation. Finally, LoRP focuses on training-free pruning without post-pruning recovery. Combining LoRP with lightweight adaptation may further improve performance under aggressive pruning budgets.

\section*{Ethical considerations}
LoRP is designed for improving the inference efficiency of large language models through training-free depth pruning. Our work does not introduce new datasets, collect human data, or involve human subjects. Since model compression may affect downstream model behavior, careful evaluation is important before deployment in safety-critical settings.

\section*{Acknowledgments}
This work was supported by Institute of Information \& communications Technology Planning \& Evaluation (IITP) under the Artificial Intelligence Innovation Human Resources Development (IITP-2026-RS-2026-25548560) grant funded by the Korea government(MSIT). This work was supported by Institute of Information \& communications Technology Planning \& Evaluation(IITP) under the Leading Generative AI Human Resources Development(IITP-2026-RS-2026-25544527) grant funded by the Korea government(MSIT). This work was supported by INHA UNIVERSITY Research Grant.


\bibliography{custom}

\newpage
\clearpage
\appendix

\section*{Appendix}
\label{sec:appendix}

\section{Inference Efficiency of Pruned LLMs}
\label{app:cost}

We complement the language modeling and downstream evaluation results with an inference-efficiency analysis on Qwen3-14B. The goal is to verify that the parameter reduction obtained through depth pruning translates into proportional reductions in runtime and GPU memory while maintaining reasonable language modeling and downstream performance.

\subsection{Benchmark Setup}
\label{app:cost-setup}

All measurements are conducted on a single NVIDIA A40 48GB GPU using CUDA 12.1, PyTorch 2.5.1, and Hugging Face Transformers in fp16 precision. We evaluate Qwen3-14B, the largest model in our benchmark suite, under pruning budgets $L_p \in \{11,13,15\}$ out of $L_t=40$ Transformer blocks.

The same pruned checkpoints used in the main perplexity and downstream evaluation experiments are used for benchmarking. Layer removal is implemented by directly deleting the corresponding Transformer blocks from \texttt{model.model.layers} and updating \texttt{model.config.num\_hidden\_layers}, resulting in a physically shallower Transformer stack without recovery modules or additional parameters.

For all configurations, inference is measured using a synthetic input of shape $(1,2048)$ with \texttt{use\_cache=False} to isolate prefill computation cost. Each configuration is evaluated using 3 warm-up forward passes followed by 20 timed forward passes with CUDA synchronization enabled before and after execution. Reported latency corresponds to the mean $\pm$ standard deviation across timed runs. Peak GPU memory is measured using \texttt{torch.cuda.max\_memory\_allocated()} during inference.

\begin{table}[h]
\centering
\small
\setlength{\tabcolsep}{4pt}
\begin{adjustbox}{width=\columnwidth}
\begin{tabular}{l c c c c}
\toprule
Config & Layers & Params & Latency (ms) & Speedup \\
\midrule
Dense                           & 40 & 14.77\,B & 676.7 $\pm$ 2.5 & 1.00$\times$ \\
\midrule
LoRP, $L_p=11$                 & 29 & 11.13\,B & 497.4 $\pm$ 1.5 & \textbf{1.36$\times$} \\
LoRP, $L_p=13$                 & 27 & 10.47\,B & 465.3 $\pm$ 1.4 & \textbf{1.45$\times$} \\
LoRP, $L_p=15$                 & 25 & 9.81\,B  & 432.8 $\pm$ 1.1 & \textbf{1.56$\times$} \\
\bottomrule
\end{tabular}
\end{adjustbox}
\caption{
Inference latency of LoRP on Qwen3-14B measured on a single NVIDIA A40 GPU (fp16, batch size 1, sequence length 2048).
}
\label{tab:latency-qwen14b}
\end{table}

\subsection{Latency Analysis}
\label{app:latency}

Table~\ref{tab:latency-qwen14b} shows that depth pruning yields near-linear reductions in inference latency. As the pruning ratio increases from 27.5\% to 37.5\%, latency decreases from 676.7\,ms to 432.8\,ms, corresponding to a 1.56$\times$ speedup over the dense baseline.

The reduction in latency closely follows the reduction in Transformer depth, indicating that training-free depth pruning translates directly into practical inference-time acceleration without additional runtime optimization.

\subsection{Memory and Performance Trade-off}
\label{app:memory}

\begin{table}[h]
\centering
\small
\setlength{\tabcolsep}{4pt}
\begin{adjustbox}{width=\columnwidth}
\begin{tabular}{l c c c c}
\toprule
Config & Peak mem & Reduction & PPL $\downarrow$ & DS $\uparrow$ \\
\midrule
Dense                           & 28.12\,GiB & -- & 12.14 & 71.22\% \\
\midrule
LoRP, $L_p=11$                 & 21.35\,GiB & $-$24.1\% & 37.44  & 52.84\% \\
LoRP, $L_p=13$                 & 20.12\,GiB & $-$28.4\% & 76.69  & 49.50\% \\
LoRP, $L_p=15$                 & 18.89\,GiB & $-$32.8\% & 204.49 & 48.07\% \\
\bottomrule
\end{tabular}
\end{adjustbox}
\caption{
Peak GPU memory and downstream performance of LoRP on Qwen3-14B.
PPL is the mean over WikiText-2 / C4 / PTB; DS is the mean zero-shot accuracy over nine commonsense reasoning tasks.
}
\label{tab:memory-qwen14b}
\end{table}

Table~\ref{tab:memory-qwen14b} shows that GPU memory usage decreases consistently as the pruning budget increases. At $L_p=15$, peak GPU memory decreases from 28.12\,GiB to 18.89\,GiB, reducing the memory requirement by 32.8\%.

The results also illustrate the trade-off between efficiency and performance under increasingly aggressive pruning budgets. Moderate pruning budgets such as $L_p=11$ maintain relatively stronger perplexity and downstream accuracy while already achieving substantial reductions in memory and latency. More aggressive pruning further improves efficiency, although with larger degradation in language modeling and downstream performance.

Overall, the results demonstrate that training-free depth pruning provides practical inference-time benefits beyond parameter reduction alone, translating directly into lower runtime and memory usage on standard hardware.

\section{Licenses and Terms of Use}
\label{app:licenses}

All datasets and models used in this work are publicly available and used in 
compliance with their respective licenses and terms of use. We use them solely 
for non-commercial research purposes, consistent with their intended use.

\paragraph{Models.} 
Qwen3-8B and Qwen3-14B~\cite{yang2025qwen3} are released under the Apache 2.0 
License. OLMo-3-7B~\cite{olmo2025olmo2} is released under the Apache 2.0 
License. Mistral-Nemo-12B~\cite{mistral2024nemo} is jointly released by Mistral 
AI and NVIDIA under the Apache 2.0 License. LLaMA-3.1-8B~\cite{dubey2024llama3} 
is released under the Llama 3.1 Community License Agreement.

\paragraph{Evaluation datasets.} 
ARC-Easy and ARC-Challenge~\cite{clark2018arc} are released under the CC BY-SA 
4.0 License. HellaSwag~\cite{zellers2019hellaswag} is released under the MIT 
License. WinoGrande~\cite{sakaguchi2021winogrande} is released under the 
CC BY 4.0 License. BoolQ~\cite{clark2019boolq} is released under the CC BY-SA 
3.0 License. OpenBookQA~\cite{mihaylov2018openbookqa} is released under the 
Apache 2.0 License. RTE~\cite{rte} is distributed for research use as part of 
the PASCAL RTE Challenge. COPA~\cite{roemmele2011copa} is publicly distributed 
by the original authors for research use in commonsense reasoning. 
RACE~\cite{lai2017race} is distributed for non-commercial research purposes 
only, as specified by its creators.

\paragraph{Calibration and perplexity datasets.} 
C4~\cite{raffel2020c4} is released under the ODC-BY License, with the 
underlying Common Crawl content also subject to Common Crawl's terms of use. 
WikiText-2~\cite{merity2017wikitext} is released under the CC BY-SA 3.0 
License (and additionally under the GFDL). The Penn 
Treebank~\cite{marcus1993ptb} is distributed by the Linguistic Data Consortium 
(LDC) under its standard research license.

\paragraph{Software.} 
The \texttt{lm-evaluation-harness}~\cite{gao2024lmeval} is released under the 
MIT License. Hugging Face Transformers and PyTorch are released under the 
Apache 2.0 and BSD-style licenses, respectively. All software is used 
consistent with its intended research use.

\end{document}